\pdfoutput=1

\documentclass[11pt]{article}

\usepackage[final]{acl}
\usepackage{xcolor}
\usepackage{graphicx}
\usepackage{colortbl}
\usepackage{adjustbox}
\usepackage{times}
\usepackage{booktabs}
\usepackage{multirow}
\usepackage{latexsym}
\usepackage[T1]{fontenc}
\usepackage{rotating} 
\usepackage{pifont}   
\usepackage[utf8]{inputenc}
\usepackage{microtype}

\newcommand{\best}[1]{\textbf{#1}} 
\newcommand{\secondbest}[1]{\underline{#1}} 
\usepackage{inconsolata}
\usepackage{tabularx}
\usepackage{graphicx}
\usepackage{subcaption} 
\usepackage{colortbl}
\usepackage{amsfonts}
\usepackage{amsmath}

\usepackage{authblk}
\usepackage[normalem]{ulem}
\newcommand{\shortname}{LLaVA-Med}
\title{Med-MoE: Mixture of Domain-Specific Experts for Lightweight Medical Vision-Language Models}
\author[1]{Songtao Jiang\textsuperscript{*}}
\author[1]{Tuo Zheng\textsuperscript{*}}
\author[2]{Yan Zhang}
\author[2]{Yeying Jin}
\author[3]{Li Yuan}
\author[1]{Zuozhu Liu}

\affil[1]{Zhejiang University}
\affil[2]{National University of Singapore}
\affil[3]{Peking University}

\begin{document}
\maketitle
\begin{abstract}


Recent advancements in general-purpose or domain-specific multimodal large language models (LLMs) have witnessed remarkable progress for medical decision-making. However, they are designated for specific classification or generative tasks, and require model training or finetuning on large-scale datasets with sizeable parameters and tremendous computing, hindering their clinical utility across diverse resource-constrained scenarios in practice. In this paper, we propose a novel and lightweight framework Med-MoE (Mixture-of-Experts) that tackles both discriminative and generative multimodal medical tasks. The learning of Med-MoE consists of three steps: multimodal medical alignment, instruction tuning and routing, and domain-specific MoE tuning. After aligning multimodal medical images with LLM tokens, we then enable the model for different multimodal medical tasks with instruction tuning, 
together with a trainable router tailored for expert selection across input modalities. Finally, the model is tuned by integrating the router with multiple domain-specific experts, which are selectively activated and further empowered by meta expert. Comprehensive experiments on both open- and close-end medical question answering (Med-VQA) and image classification tasks across datasets such as VQA-RAD, SLAKE and Path-VQA demonstrate that our model can achieve performance superior to or on par with state-of-the-art baselines, while only requiring approximately 30\%-50\% of activated model parameters. Extensive analysis and ablations corroborate the effectiveness and practical utility of our method. Our code is released at \href{https://github.com/jiangsongtao/Med-MoE}{\texttt{https://github.com/jiangsongtao/Med-MoE}}.

\end{abstract}

\section{Introduction}

Creating systems with human-level multimodal understanding is essential for medical decision-making ~\citep{miao2022research,goyal2016towards,de2023visual,antol2015vqa}. Recent progress on Multimodal Large Language Models (MLLMs) such as LLaVA~\citep{liu2024visual}, MiniGPT4-V2~\citep{chen2023minigptv2}, CogVLM~\citep{wang2023cogvlm} have demonstrated great performance across multimodal tasks, however, they are less effective in the medical domain as they are usually trained with web contents which differ significantly from the medical data. Domain-specific models such as Med-Flamingo~\citep{moor2023med}, Med-PaLM M~\citep{singhal2023towards}, and LLaVA-Med~\citep{li2024llava} exhibit promising results across various medical tasks, such as medical visual question-answering (Med-VQA), by training with medical domain data. However, these models are usually tailed for certain kinds of tasks, such as close- or open-end VQA, while in practice, medical MLLMs need to handle both discriminative and generative tasks to provide more reliable and interpretable decisions. Moreover, existing models are usually obtained with heavy LLMs with sizeable parameters, such as LLama(7B) in LLaVA-Med, leading to high training and inference costs and hindering its practical utility to broad clinical practitioners.  

\begin{figure}[!t]
    \centering
    \includegraphics[width=1\linewidth]{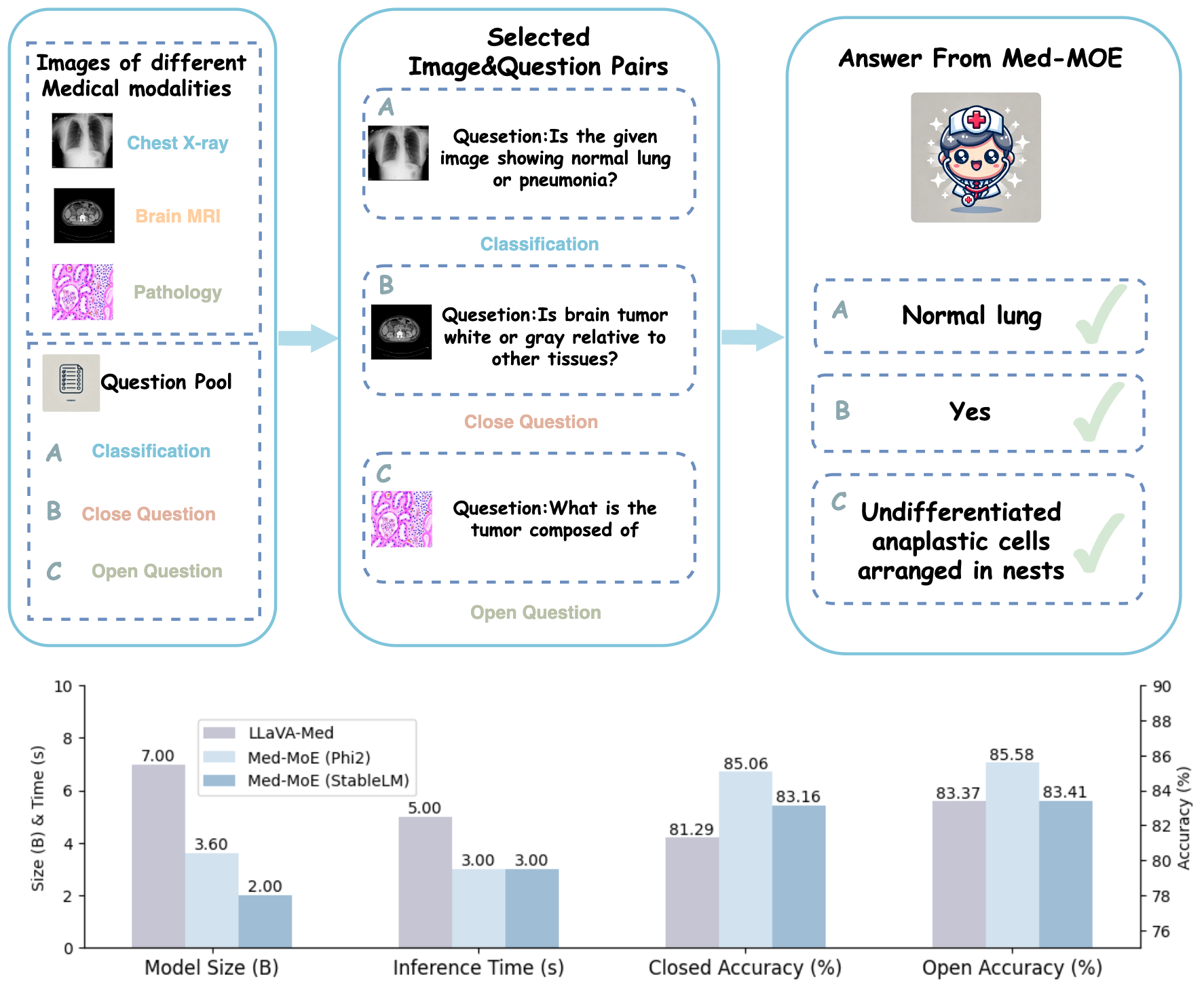}
\caption{
\textbf{Upper:} This figure showcases our model's capability in addressing three primary types of Medical VQA challenges and image classification tasks.
\textbf{Lower:} Comparison between Med-MoE and LLaVA-Med, emphasizing Med-MoE's advantages in inference speed, model size, and its superior performance.}

    \label{fig:case1}
\end{figure}

\begin{figure*}[!ht]
    \centering
    \includegraphics[width=1\linewidth]{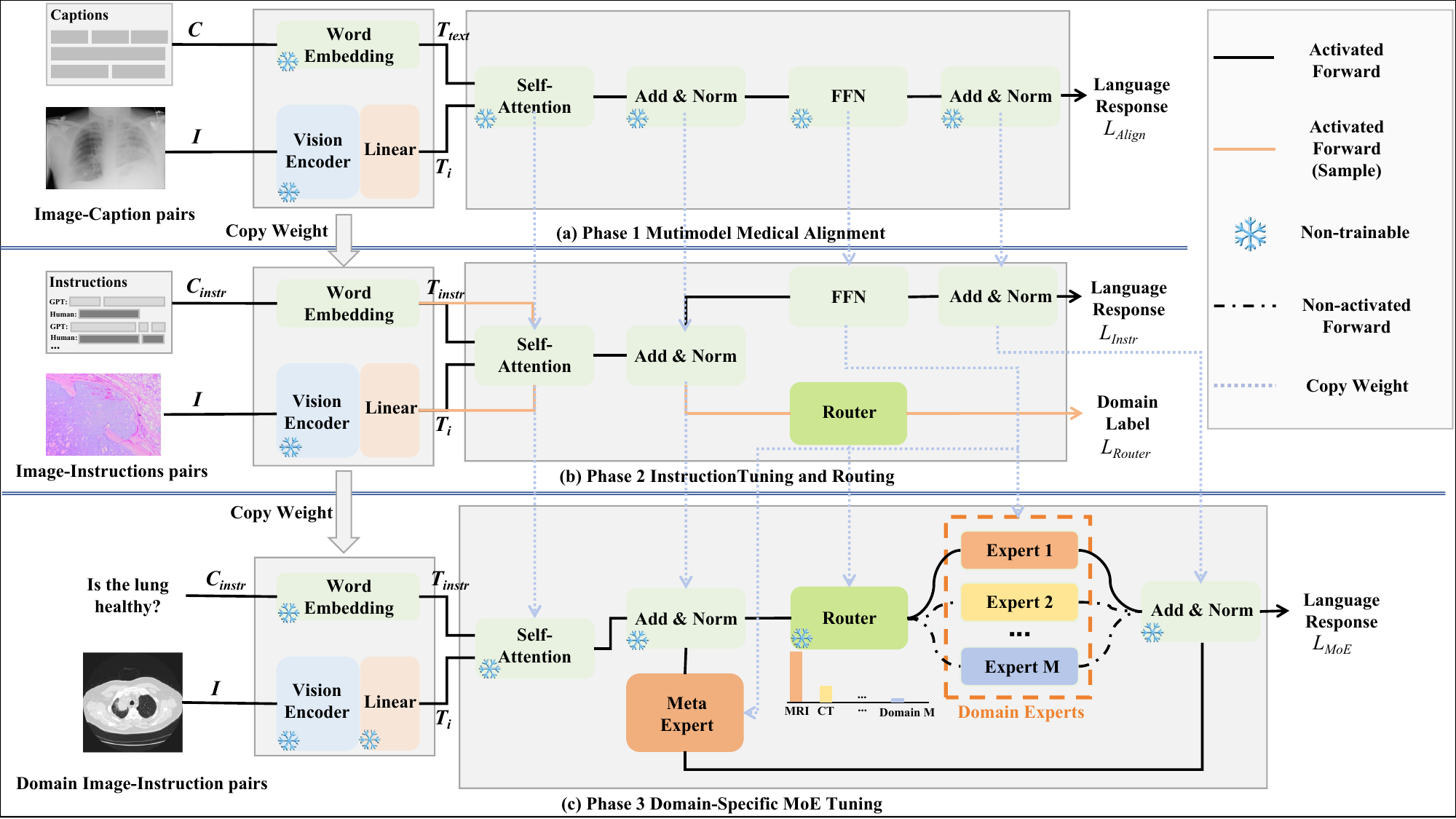}
\caption{The framework of Med-MoE with three phases. 
}
    \label{fig:pipeline}
\end{figure*}

It is appealing but challenging to build lightweight yet effective medical MLLMs for multimodal decision-making ~\citep{petersson2022challenges,kelly2019key,liao2024lightm}. 
Recent research shows that scaling up the quantity or quality of training data, as well as increasing the size of the model, can result in enhanced performance ~\citep{gao2024sphinx,shi2024we,xue2024repeat,shen2023slimpajama,lu2023empirical}. However, training and deploying these models also demand substantial computational resources, rendering them less appealing for numerous clinical practitioners who may lack sufficient computing power ~\citep{lu2023empirical,crawford2021atlas,thompson2020computational}. For instance, many institutions may not possess powerful GPUs such as NVIDIA A100 cards to tune LLama-7B model family, e.g., LLaVA-Med. Moreover, the medical data differs drastically from web contents, and its inherent multi-modality, such as imaging from CT, MRI, X-ray and pathology, presents additional challenges to develop effective yet lightweight medical MLLMs~\cite{acosta2022multimodal,xu2024comprehensive}. This task becomes even more difficult when considering the requirements for reliability and interpretability in decision-making~\cite{salahuddin2022transparency,vellido2020importance}.  

Recent work explores cost-effective training of light-weight LLMs by architecture design, training procedure or hardware optimization etc~\cite{dubiel2024device,zhao2024galore,hu2024bliva,zhou2024tinyllava}. Among these techniques, the Mixture-of-Expert (MoE) strategy has shown great potential for general-purpose training ~\citep{chen2022towards,he2021fastmoe,jacobs1991adaptive,eigen2013learning}, e.g., the Mixtral family employs sparse MoEs to achieve competing performance with LLama-70B with only 12.9B active parameters~\cite{jiang2024mixtral}; the MoE-LLaVA propose a MoE-based sparse large VLM framework with novel training strategies~\cite{lin2024moe}. By combining multiple small-scale sub-modules, i.e., experts, and activating only the top-$k$ relevant experts for each task, the MoE model can achieve good performance with much less computing cost. Despite these successes in general domains, current MoEs often overlook the specialization and synergy of experts required in medical contexts, and their applicability in the medical domain remains unexplored.  

In this paper, we propose a lightweight and effective framework Med-MoE for multimodal generative or discriminative Med-VQA and classification tasks. Our Med-MoE incorporates multiple domain-specific experts along with global meta expert, emulating the workflow in hospitals where various departments collaborate together for disease diagnosis. In particular, the Med-MoE takes lightweight LLMs with smaller sizes of parameters as the base model of experts, which are first trained with medical image and caption pairs to align visual and textual modalities. Afterward, the model is trained with medical instruction following datasets to better perform multimodal medical tasks. Meanwhile, a router is trained to identify different medical image modalities, enabling better selection across multiple domain-specific experts during decision-making. Inspired by the well-known ResNet~\cite{he2016deep} architecture and the Multi-Disciplinary Team (MDT) diagnosis mechanism in clinics, we propose to add an additional meta expert in the shortcut, as shown in Figure~\ref{fig:pipeline}, which captures global medical information to assist the specified expert for better performance. During inference, only the meta expert and the selected experts are activated, leading to a lightweight model with only a small portion of activated parameters.

The Med-MoE consistently demonstrates significant performance improvements across diverse medical datasets, encompassing both open- and close-end Med-VQA and medical image classification tasks in VQA-RAD, SLAKE, PathVQA, PneumoniaMNIST, and OrganCMNIST. Comprehensive experiments show that our Med-MoEs, which are constructed with two small-scale LLMs, i.e., Phi2 (2.7B)~\cite{abdin2024phi} and StableLM (1.7B)~\cite{bellagente2024stable}, can attain performance superior to or on par with the state-of-the-art LLaVA-Med (7B) model, with only 2.0-3.6B activated parameters. Extensive ablations and analysis demonstrate the efficacy of our Med-MoE in advancing multimodal medical tasks and highlight its potential to enhance outcomes in resource-limited healthcare settings. 

\section{Methods}
The training of Med-MoE includes three phases, as illustrated in Figure \ref{fig:pipeline}. First, we perform multimodal medical alignment to help the LLM to comprehend medical images by leveraging the vision encoder's image tokens. Next, we conduct instruction tuning to enable the model to execute various medical tasks and enhance its instruction-following ability. Meanwhile, a router is trained with a small amount of labeled data to characterize the input modality. Finally, we perform domain-specific MoE tuning by replacing the model's FFN with sparsely activated experts, where a meta-expert is always activated to capture global information. 

\subsection{Phase 1: Multimodal Medical Alignment}
In this phase, we train only the MLP following the vision encoder to achieve modality alignment. We align visual and textual modalities by curating a dataset of medical images \( I \) paired with corresponding captions \( C \). The images \( I \) are fed into a vision encoder \( E_v \) to produce image tokens \( T_i \) (\( T_i = E_v(I_i) \)), and the captions \( C \) are tokenized into text tokens \( T_{\text{text}} \) (\( T_{\text{text}} = \text{Tokenizer}(C_i) \)). The concatenated tokens \( T_{\text{comb}} \) are fed into the LLM, which is trained to generate the continuation of text tokens, minimizing the self-supervised loss:
\begin{equation}
\small 
\mathcal{L}_{\text{Align}} = - \sum_{i=1}^{N} \log p \left( T_{\text{text}}^{[P+i]} \mid T_{\text{comb}}, T_{\text{text}}^{[:i-1]} \right).
\end{equation}

\subsection{Phase 2: Instruction Tuning and Routing}


This phase aims to enhance the model's ability to follow complex medical instructions to perform various multimodal tasks and train a router for expert selection in the next phase. The instruction tokens \( T_{\text{instr}} \) and image tokens \( T_i \) are concatenated into \( T_{\text{comb}} \) and fed into the LLM. The model is trained using a dataset of medical queries and responses to generate accurate responses, minimizing the loss:
\begin{equation}
\small
\mathcal{L}_{\text{Instr}} = - \sum_{i=1}^{N} \log p \left( T_{\text{resp}}^{[P+i]} \mid T_{\text{comb}}, T_{\text{resp}}^{[:i-1]} \right). 
\end{equation}

We also train a router to predict the input modality using a small subset of data with the loss:
\begin{equation}
\small
\mathcal{L}_{\text{Router}} = - \sum_{i=1}^{M} y_i \log p (y_i \mid T_{\text{comb}}),
\end{equation}
where \( y_i \) is the true label of the input image modality, i.e., CT, MRI, Pathology and X-ray, etc. In this phase, the vision encoder is frozen, while all other components are trained.

\subsection{Phase 3: Domain-Specific MoE Tuning}
Finally, we replace the LLM's FFN with MoE(mixture-of-experts) architecture. The router, trained in phase 2, assigns inputs to specific experts, while a meta-expert is always activated to capture global information. The MoE layer's output is a weighted combination of the experts' outputs. The domain-specific experts and the meta-expert are initialized with the FFN weights from the model trained in phase 2. In this phase, the router from phase 2 is used and frozen, so only the domain-specific experts and the meta-expert are trained.
\begin{equation}
\small
\mathbf{O}_{\text{MoE}} = \sum_{i=1}^{K} G_i E_i + E_{\text{meta}},
\end{equation}
where \( G_i \) is the gating function provided by the router, \( E_i \) are the domain-specific experts, and \( E_{\text{meta}} \) is the meta-expert. The training loss is:
\begin{equation}
\small
\mathcal{L}_{\text{MoE}} = - \sum_{i=1}^{N} \log p \left( T_{\text{resp}}^{[P+i]} \mid \mathbf{O}_{\text{MoE}}, T_{\text{resp}}^{[:i-1]} \right).
\end{equation}
In the end, the model is fine-tuned for specific medical domains, leveraging expert knowledge to provide highly accurate and relevant responses across the open- and close-end and classification tasks. 

\section{Experiment}

\subsection{Experiment Settings}
\noindent \textbf{Dataset:} We utilize well-organized datasets provided by LLaVA-Med~\citep{li2024llava} for alignment and instruction tuning in phase 1\&2, see details in Supplementary Figure \ref{tab:data_summary}. 
In MoE-tuning phase, we employ VQA-RAD~\citep{lau2018dataset}, SLAKE~\citep{liu2021slake},  PathVQA~\citep{he2020pathvqa} with open- and close-end QA pairs for Med-VQA tuning and evaluation. For classification task, we use the PneumoniaMNIST and OrganCMNIST from ~\citep{yang2023medmnist}. Detailed information and examples of Med-MoE's responses are shown in Supplementary. 

\noindent \textbf{Evaluation Metrics:} We employ the accuracy for closed-set questions and recall for open-set questions, being consistent with existing work like LLaVA-Med for a fair comparison. In Table \ref{tab:open_acc}, we also evaluate the exact match and BLEU scores for comprehensive evaluation.  

\noindent \textbf{Experiment Setup:} We select two small LLMs, i.e., StableLM (1.7B) and Phi2 (2.7B), as the base model, see Figure \ref{tab:model_arichi}. The size of activated parameters in resulting Med-MoEs are 2.0B (Med-MoE StableLM) and 3.6B (Med-MoE Phi2), with additional parameters from domain-specific experts and meta-experts. To ensure a fair comparison with LLaVA-Med, we also train a LLaVA-Med model using the Phi2 (2.7B) backbone. This allows us to compare the performance under the same LLM backbone. We also investigate the versatility of our method by combining it with other cost-efficient approaches, such as LoRA-based methods. Our experimental hyperparameters are shown in Supplementary Table \ref{tab:setting}.

\noindent \textbf{Baselines:} We compare our method with a diverse set of baselines: (1) \textbf{CLIP-based methods}, such as BiomedCLIP and CLIP-ViT~\citep{zhang2023large,eslami2023pubmedclip}, which are state-of-the-art in this category but are limited by their reliance on candidate words for answering questions in open settings; (2) \textbf{OFA (One for All)-based models}, like the recent BiomedGPT~\citep{zhang2023biomedgpt}, which leverage generative multimodal pretraining and have shown promising performance in the medical field, but their lack of multi-turn dialogue capability, due to not being LLM-based, restricts their usage in clinical practice; (3) \textbf{MLLM-based models}, including Med-Flamingo and the state-of-the-art LLaVA-Med, which, despite their impressive VQA performance, have large parameter sizes (7B and above) that hinder their applicability in real-world clinical settings. In classification tasks, we compare with ViT-based methods and the latest Med-Mamba~\citep{yue2024medmamba}. Notably, our Med-MoE, an MLLM-based method, offers multi-turn dialogue capabilities for open VQA settings which are not present in traditional methods, while exhibiting effective training/inference and competing performance.

\begin{table*}[ht!]
\centering
\resizebox{\textwidth}{!}{
\begin{tabular}{ll|cc|cc|cc|c}  
 & & \multicolumn{2}{c|}{\bf VQA-RAD} & \multicolumn{2}{c|}{\bf SLAKE} & \multicolumn{2}{c|}{\bf PathVQA} & \multicolumn{1}{c}{\bf Act.} \\
Method  &  & Open   & Closed    & Open   & Closed  & Open &  Closed & \\
\hline
\multicolumn{9}{l}{\it Representative \& SoTA methods with numbers reported in the literature (Non-MLLM Based Methods) } \\  
\hline
VL Encoder–Decoder~\citep{bazi2023vision} & & - & 82.47 & - & - & - & 85.61 & - \\
Q2ATransformer~\citep{liu2023q2atransformer} & & - & 81.20 & - & - & 54.85 & 88.85 & - \\
Prefix T. Medical LM~\citep{van2023open} & & - & - & - & 82.01 & - & 87.00 & - \\
PubMedCLIP~\citep{eslami2023pubmedclip} & & - & 80.00 & - & 82.50 & - & - & - \\
BiomedCLIP~\citep{zhang2023large} & & - & 79.80 & - & 89.70 & - & - & - \\
M2I2~\citep{li2022self} & & - & 83.50 & - & 91.10 & - & 88.00 & - \\
BiomedGPT-S~\citep{zhang2023biomedgpt} & & 13.40 & 57.80  & 66.50 & 73.30 & 10.70 & 84.20 & - \\
BiomedGPT-M~\citep{zhang2023biomedgpt}& & 53.60 & 65.07 & 78.30 & 86.80 & 12.5 & 85.70 & - \\
CLIP-ViT w/ GPT2-XL& & - & - & 84.30 & 82.10 & 40.0 & 87.00 & - \\
\hline
\multicolumn{9}{l}{\it Supervised finetuning results (MLLM Based Methods)} \\
\hline
LLaVA & & 50.00 & 65.07 & 78.18 & 63.22 & 7.74 & 63.20 & 7B \\
\shortname{} (LLama7B) & & \underline{61.52} & {\bf 84.19} & 83.08 & \underline{85.34}  &  \underline{37.95}   & {91.21} & 7B \\
\shortname{} (Vicuna7B) & & \bf 64.39 & 81.98 & \underline{84.71} & 83.17  &  \bf 38.87  & \underline{91.65} & 7B \\
\shortname{} (Phi2.7B) &  & 54.83 & {81.35} & 81.29  &  83.29  & {31.73} & 90.17 & 2.7B \\
\rowcolor[HTML]{CBC5D3} Med-MoE (Phi2) & & {58.55} & \underline{82.72} &  \bf 85.06 & \bf 85.58 & 34.74 & \bf 91.98 & 3.6B \\
\rowcolor[HTML]{CBC5D3} Med-MoE (StableLM) & & 50.08 & 80.07 &  {83.16} & {83.41} &  33.79 & 91.30 & 2.0B \\

\hline
\hline
\multicolumn{9}{l}{\it Zero-shot results} \\
\hline
LLaVA-Med (LLama7B) & & \underline{36.23} & 60.16 & \underline{41.72} & 47.60 & \bf 10.86 & 59.75 & - \\
\rowcolor[HTML]{CBC5D3} Med-MoE (Phi2) & & \bf 36.73 & \underline{61.75} & \bf 43.93 & \bf 56.97 &  6.94 & \underline{66.46} & - \\
\rowcolor[HTML]{CBC5D3} Med-MoE (StableLM) & & 28.02 & \bf{66.91} & 40.63 & \underline{52.64} &   \underline{9.40} & \bf{69.09} & - \\
\hline
\end{tabular}
}
\caption{Performance on Med-VQA tasks. \textbf{Bold} denotes the best performance; \underline{underlined} denotes the second-best.}
\label{tab:main_table}
\end{table*}

\begin{table*}[ht!]
\centering
\resizebox{\textwidth}{!}{ 
\begin{tabular}{ll|ccc|ccc|ccc}  
 & & \multicolumn{3}{c|}{\bf VQA-RAD} & \multicolumn{3}{c|}{\bf SLAKE} & \multicolumn{3}{c}{\bf PathVQA} \\
\textbf{Method}  &  & \textbf{EMS} & \textbf{R} & \textbf{BS} & \textbf{EMS} & \textbf{R} & \textbf{BS} & \textbf{EMS} & \textbf{R} & \textbf{BS} \\
\hline
\rowcolor[HTML]{EFEFEF} \shortname{}  & 7B & \underline{58.33} & \bf61.52 & \underline{54.13} & \underline{82.83} & 83.08 & \underline{81.69} & \bf37.95 & \bf36.86 & \underline{32.89}  \\
\rowcolor[HTML]{EFEFEF} Med-Flamingo(Few-Shot)~\citep{moor2023med}  & 9B & 20.00 & - & - & - & - & - & 31.00& - &  -  \\
\rowcolor[HTML]{EFEFEF} PaLM-E~\citep{tu2024towards}  & 84B & - & - & \bf59.19 & - & - & 52.65 & - & - &   \bf54.92  \\
\rowcolor[HTML]{CBC5D3} Med-MoE (Phi2) & 3.6B & \bf59.69 & \underline{58.55} & 52.95 & \bf84.46 & \bf85.06 & \bf83.16 & \underline{34.37} & \underline{34.74} & 32.85  \\
\rowcolor[HTML]{CBC5D3} Med-MoE (StableLM)  & 2.0B & 52.53 & 50.08 & 45.67 & 82.44 & \underline{83.16} & 81.53 & 33.60 &33.79 & 32.67  \\
\hline
\end{tabular}
}
\caption{Detailed comparison regarding more metrics (Supplementary \ref{sec:gongshi}) in Open settings. 
}
\label{tab:open_acc}
\end{table*}

\begin{table}[!ht]
\centering
\setlength{\tabcolsep}{3pt}
\resizebox{0.5\textwidth}{!}{
    \begin{tabular}{lcc}
        \toprule
        \textbf{Methods} & \textbf{PneumoniaMNIST} & \textbf{OrganCMNIST} \\
        \midrule
        Med-Mamba~\citep{yue2024medmamba} & \underline{91.2} & \textbf{92.4} \\
        AutoKeras~\citep{jin2019auto} & 87.8 & 87.9 \\
        BiomedGPT & 90.8 & 88.9 \\
        Med-MoE (StableLM) & 89.3 & 88.6 \\
        Med-MoE (Phi2) & \textbf{91.4} & \underline{89.9} \\
        \bottomrule
    \end{tabular}
    }
\caption{Image classification accuracy comparison. 
}
\label{tab:classification}
\end{table}

\subsection{Main Results}
\noindent \textbf{Zero-shot Performance on Med-VQA tasks:} Our models exhibit notable improvements in zero-shot performance across various medical VQA tasks. The Med-MoE (Phi2) model boosts scores by approximately 1.4\% in VQA-RAD Open, 2.6\% in VQA-RAD Closed, 5.3\% in SLAKE Open, and 9.4\% in SLAKE Closed compared to LLaVA-Med (LLama7B). The Med-MoE (StableLM) variant achieves around 6.8\% higher in VQA-RAD Closed, 5.0\% in SLAKE Closed, and 9.3\% in PathVQA Closed, demonstrating robust performance. These results highlight the superior effectiveness of Med-MoE models in zero-shot settings.

\noindent \textbf{Comparison with SOTA Methods on Med-VQA:} Overall, Med-MoE can achieve superior or competing performance with the best-performing LLaVA-Med (7B) with only 2.0B or 3.6B activated parameters. In particular, Med-MoE (Phi2) surpasses the best LLaVA-Med variants in SLAKE Open (85.06), SLAKE Closed (85.58), and PathVQA Closed (91.98), and shows competing performance on the rest tasks. Med-MoE (StableLM) also exhibits better performance than the LLaVA-Med (Phi-2.7B) in most scenarios, with only 2.0B activated parameters. Its performance is also on par with LLaVA-Med (7B) in many scenarios, with even better performance in SLAKE Closed. These results highlight the effectiveness and strong potential of Med-MoE to establish new benchmarks across various datasets and tasks.

\noindent \textbf{Results on Medical Image Classification:} In contrast to most existing LLM-based work, e.g., LLaVA-Med, which only evaluates the performance on Med-VQA, we further evaluate Med-MoE on classification tasks for comprehensive analysis. As shown in Table \ref{tab:classification}, Med-MoE (Phi2) achieves 91.4\% accuracy on PneumoniaMNIST, outperforming BiomedGPT and Med-Mamba. It also showcases the second-best performance on OrganCMNIST, closely following Med-Mamba. The performance of Med-MoE (StableLM) is a little bit worse. Overall, the classification performance of Med-MoE is quite promising, while its performance might be boosted if more relevant data rather than image-caption pairs could be used for model alignment and tuning in the initial phases. 

\section{Ablation and Analysis}

\begin{table}[ht!]
\centering
\scriptsize
\resizebox{0.5\textwidth}{!}{ 
\setlength{\tabcolsep}{10pt}
\begin{tabular}{l|c|c}
\hline
\textbf{Method} & \textbf{SFT} & \textbf{MoE-Tuning} \\
\hline
{VQA-RAD (Open)} & 54.83 & \textbf{58.55 (+3.72)} \\
{VQA-RAD (Closed)} & 81.35 & \textbf{82.72 (+1.37)} \\
{SLAKE (Open)} & 81.29 & \textbf{85.06 (+3.77)} \\
{SLAKE (Closed)} & 83.29 & \textbf{85.58 (+2.29)} \\
{PathVQA (Open)} & 31.73 & \textbf{34.74 (+3.01)} \\
{PathVQA (Closed)} & 90.17 & \textbf{91.98 (+1.81)} \\
\hline
\end{tabular}
}
\caption{Comparison of SFT and MoE Tuning.}
\label{tab:ablation}
\end{table}

\begin{table}[ht!]
\centering
\setlength{\tabcolsep}{5pt} 
\renewcommand{\arraystretch}{1.2} 
\resizebox{0.5\textwidth}{!}{ 
\begin{tabular}{l|c|c}
\hline
\textbf{Method} & \textbf{No Meta Expert} & \textbf{With Meta Expert} \\
\hline
{VQA-RAD (Open)} & 54.37 & \textbf{58.55 (+4.18)} \\
{VQA-RAD (Closed)} & 81.42 & \textbf{82.72 (+1.30)} \\
{SLAKE (Open)} & 81.54 & \textbf{85.06 (+3.52)} \\
{SLAKE (Closed)} & 82.45 & \textbf{85.58 (+3.13)} \\
{PathVQA (Open)} & 32.12 & \textbf{34.74 (+2.62)} \\
{PathVQA (Closed)} & 90.19 & \textbf{91.98 (+1.79)} \\
\hline
\end{tabular}
}
\caption{Ablation on the meta expert.}
\label{tab:meta_expert_comparison}
\end{table}

\begin{table}[ht!]
\centering
\setlength{\tabcolsep}{5pt} 
\renewcommand{\arraystretch}{1.2} 
\resizebox{0.5\textwidth}{!}{ 
\begin{tabular}{l|c|c}
\hline
\textbf{Method} & \textbf{Learned Router} & \textbf{Router (Ours)} \\
\hline
{VQA-RAD (Open)} & 56.33 & \textbf{58.55 (+2.22)} \\
{VQA-RAD (Closed)} & 82.19 & \textbf{82.72 (+0.53)} \\
{SLAKE (Open)} & 82.75 & \textbf{85.06 (+2.31)} \\
{SLAKE (Closed)} & 84.59 & \textbf{85.58 (+0.99)} \\
{PathVQA (Open)} & 33.40 & \textbf{34.74 (+1.34)} \\
{PathVQA (Closed)} & 91.19 & \textbf{91.98 (+0.79)} \\
\hline
\end{tabular}
}
\caption{Ablation on the routing mechanism. }
\label{tab:router_ablation}
\end{table}

\noindent\textbf{Ablation of Router:} We evaluate the effectiveness of our routing mechanism to the general MoE routing mechanism across Med-VQA datasets. Results in Table \ref{tab:router_ablation} with the Phi2.7B model show that our router achieves consistent improvements. The improvements on Open settings are even more evident than in Closed settings, demonstrating the effectiveness of our routing mechanism on more challenging Open scenarios. 


\noindent\textbf{Ablation of Meta Expert:} 
We evaluate the impact of the meta expert with ablation results with the Phi2.7B model in Table \ref{tab:meta_expert_comparison}. We can notice that the meta expert can bring consistent and significant improvements over all Med-VQA settings, with improvements of 1.30-4.18\%. These consistent improvements underscore the critical role of the meta expert in enhancing the model's ability to cope with various multimodal medical tasks.


\noindent\textbf{Ablation of Domain-Specific MoE-Tuning:} To assess the benefits of the MoE-Tuning over traditional Supervised Fine-Tuning (SFT), we conduct an ablation study with the Phi2.7B model. Results across three Med-VQA datasets (Table \ref{tab:ablation}) demonstrate that the MoE-Tuning can lead to better performance. These results demonstrate the effectiveness of our MoE architecture, giving rise to better performance than tuning a dense FFN. 

\begin{table*}[ht!]
\centering
\setlength{\tabcolsep}{5pt} 
\renewcommand{\arraystretch}{1.2} 
\resizebox{\textwidth}{!}{ 
\begin{tabular}{ll|cc|cc|cc|c|c}  
 & & \multicolumn{2}{c|}{\bf VQA-RAD} & \multicolumn{2}{c|}{\bf SLAKE} & \multicolumn{2}{c|}{\bf PathVQA} & \multicolumn{1}{c|}{\bf Act.} & \multicolumn{1}{c|}{\bf Rank} \\
Method  &  & Open   & Closed    & Open   & Closed  & Open &  Closed &  &  \\
\hline
\rowcolor[HTML]{EFEFEF}\shortname{} (LLama7B) with LoRA & & 58.22 (-3.30) & 82.13 (-2.06) & 81.29 (-1.79) & 83.37 (-1.97) & 34.33 (-3.62) & 90.12 (-1.09) & 7B & 128 \\
\rowcolor[HTML]{EFEFEF}\shortname{} (Vicuna7B) with LoRA & & 61.37 (-3.02) & 80.03 (-1.95) & 82.02 (-2.69) & 81.74 (-1.43) & 36.78 (-2.09) & 90.67 ({-0.98}) & 7B & 128 \\
\rowcolor[HTML]{CBC5D3} Med-LoRAMoE (Phi2) & & 58.12 (\best{-0.43}) & 82.35 (\best{-0.37}) & 83.58 (\best{-0.12}) & 84.85 \secondbest{(-0.49)} & 32.62 (\best{-0.63}) & 91.18 (\best{-0.80}) & 3.6B & 256 \\
\rowcolor[HTML]{CBC5D3} Med-LoRAMoE (Phi2) & & 57.20 (\secondbest{-1.35}) & 81.75 (\secondbest{-0.97}) & 83.95 (\secondbest{-0.35}) & 84.37 (-1.21) & 33.03 (\secondbest{-0.98}) & 90.83 (-1.15) & 3.6B & 128 \\
\rowcolor[HTML]{CBC5D3} Med-LoRAMoE (StableLM) & & 47.83 (-2.25) & 79.04 (-1.03) & 82.12 (-1.04) & 82.45 ({-0.96}) & 33.28 (-1.46) & 90.80 (\secondbest{-0.89}) & 2.0B & 256 \\
\rowcolor[HTML]{CBC5D3} Med-LoRAMoE (StableLM) & & 45.74 (-2.65) & 78.31 (-1.76) & 82.27 (-0.89) & 83.17 (\best{-0.24}) & 32.20 (-2.53) & 90.62 (-1.08) & 2.0B & 128 \\
\end{tabular}
}
\caption{Comparison of models with LoRA across VQA in open and closed settings. Deltas indicate performance changes compared to models without LoRA. The smallest changes are in bold while the second smallest are underlined.}
\label{tab:lora_comparison}
\end{table*}

\noindent\textbf{Comparison with LoRA-based Methods:}
We further investigate the compatibility of our methods with other lightweight techniques, i.e., LoRA. As shown in Table \ref{tab:lora_comparison}, by integrating with LoRA, Med-MoE exhibits much less performance degradation compared to LLaVA-Med. For instance, in the SLAKE Closed setting, Med-MoE (Phi2) with LoRA exhibits only a 0.49\% performance drop, whereas LLaVA-Med with LoRA experiences 1.97\% degradation. Furthermore, we observe that LoRA reduces GPU memory usage during training, and our Med-MoE requires fewer activated parameters during inference. Consequently, the integration of LoRA with Med-MoE achieves lightweight learning in terms of both training and inference. These findings indicate that Med-MoE presents an appealing practical choice for medical tasks, delivering promising performance at significantly lower computational costs.

\begin{figure}[ht!]
    \centering
    \resizebox{\columnwidth}{!}{ 
        \includegraphics{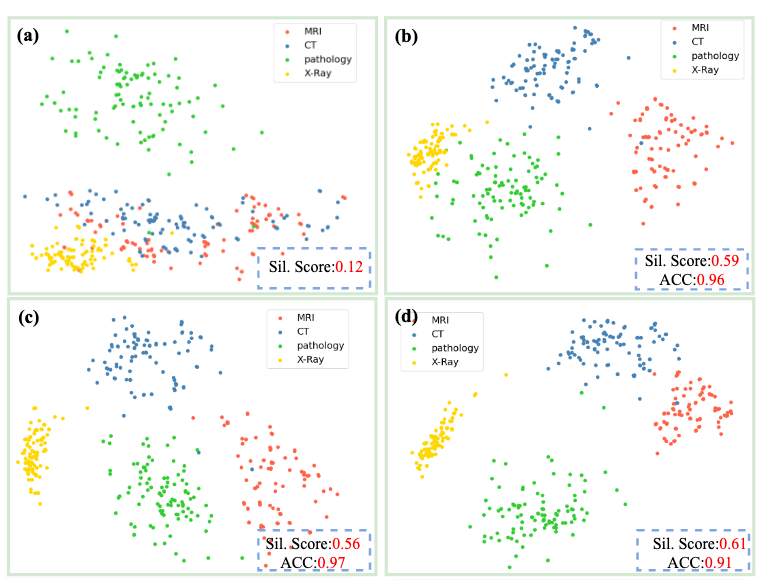}
    }
    \resizebox{\columnwidth}{!}{ 
    \setlength{\tabcolsep}{2pt} 
    \renewcommand{\arraystretch}{1.2} 
    \begin{tabular}{l|c|cc|cc|cc}
    \hline
    \textbf{Method} & \textbf{MLP Parameters} & \multicolumn{2}{c|}{\textbf{VQA-RAD}} & \multicolumn{2}{c|}{\textbf{SLAKE}} & \multicolumn{2}{c}{\textbf{PathVQA}} \\
     &  & \textbf{Open} & \textbf{Closed} & \textbf{Open} & \textbf{Closed} & \textbf{Open} & \textbf{Closed} \\
    \hline
    \textbf{a} & 0.02MB(MLP x 1) & 44.53 & 76.48 & 81.85 & 81.78 & 31.05 & 90.42 \\
    \textbf{b} & 0.02MB(MLP x 1) & \textbf{45.74} & \textbf{78.31 } & 82.27 & \textbf{83.17 } & \textbf{32.20} & 90.62 \\
    \textbf{c} & 1.00MB(MLP x 2) & 45.03 & 77.64 & \textbf{82.36 } & 82.98 & 32.13 & \textbf{90.67 } \\
    \textbf{d} & 1.13MB(MLP x 3) & 44.98 & 77.85 & 81.77 & 82.09 & 31.87 & 90.17 \\
    \hline
    \end{tabular}
    }
    \caption{Visualization of task embeddings and performance using routers under varied settings. Silhouette score (sil. score) denotes superior task differentiation. Supplementary Figure \ref{fig:router_sup} illustrates Phi2's embeddings.}
    \label{fig:routing_ablation}
\end{figure}

\noindent\textbf{Effect of Architectures and Training Data of Router:}
Figure \ref{fig:routing_ablation} investigates the effectiveness of different MLP structures in the router. We can notice that complicated MLPs, e.g., using 3 MLP layers, might not give rise to consistent improvements and may even lead to overfitting. As shown in Figure \ref{fig:routing_ablation}, a simple MLP with 1 or 2 layers can learn good embeddings of the input modality, resulting in clusters with clear boundaries, as well as high accuracy and Silhouette score. Figure 4 confirms effective modality differentiation with well-separated embeddings of image-text pairs post router processing. Moreover, we also investigate the effectiveness of our router when trained with different numbers of modality labels. Results in Figure \ref{fig:data_router} demonstrate that the training of our router only require a small set of modality labels without incurring much computational cost. 

\begin{figure*}[ht!]
    \centering
    \includegraphics[width=1\linewidth]{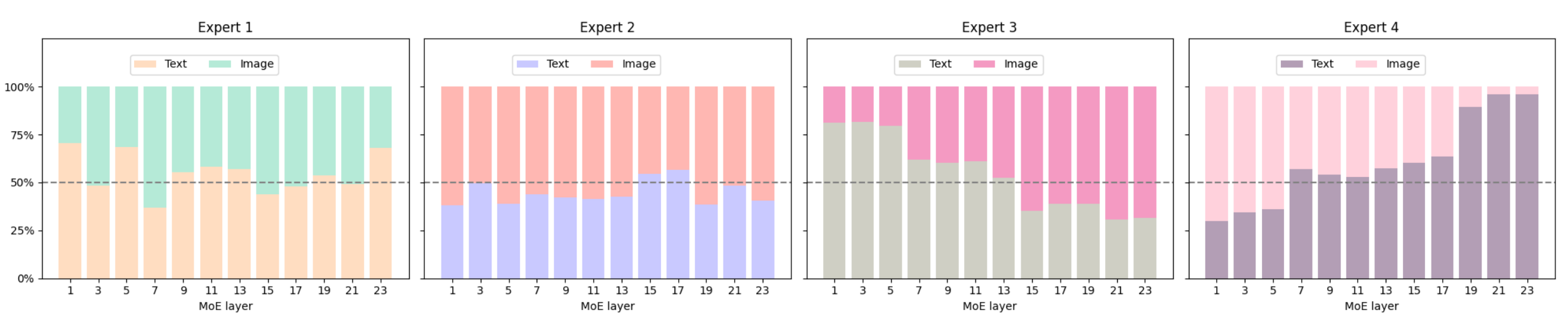}
\caption{Visualization of expert specialization in processing image and text tokens under the MRI modality. Results for other modalities are in Supplementary Figure \ref{fig:other_image_token}.}
    \label{fig:ana2_2}
\end{figure*}

\noindent\textbf{Image and Text Specialization in Experts:} As shown in Figure \ref{fig:ana2_2}, we visualize the domain specificity of experts when processing MRI inputs. Each expert shows distinct preferences for handling text or image information. For example, Expert 1 mainly handles text data, while Expert 2 has no preference for text or image data. Expert 3 focuses on image data, whereas Expert 4 specializes in text data. This differentiation highlights the Router's ability to enhance MoE model efficiency and performance by assigning tasks to suitable experts.

\begin{figure*}[!ht]
    \centering
    \includegraphics[width=1\linewidth]{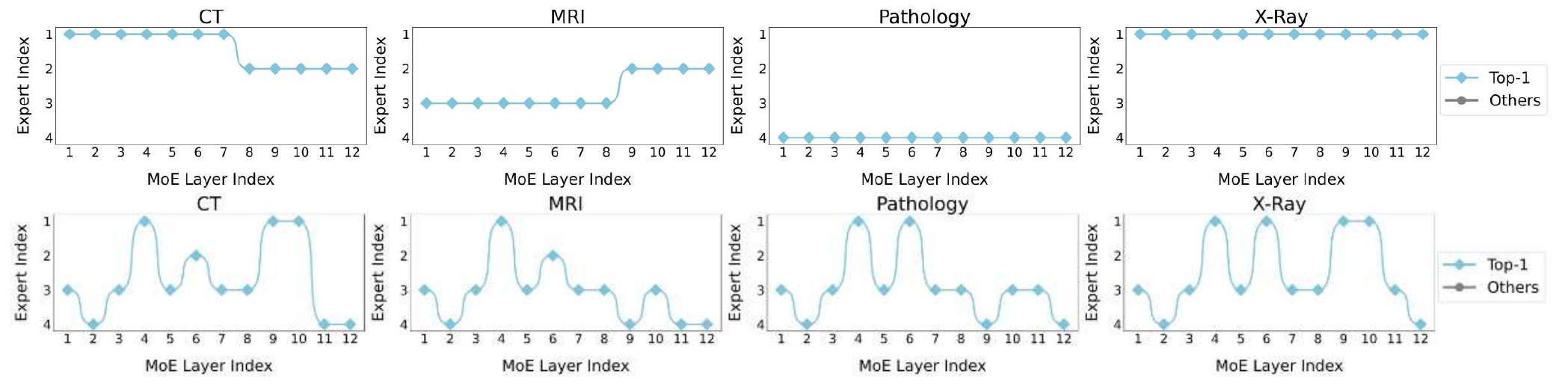}
\caption{\textbf{Upper:} Expert activations for four modalities handled by our router of Med-MoE in each MoE layer. \textbf{Lower:} Expert activations for four modalities handled by the standard learned router in each MoE layer.}
\label{fig:routing}
\end{figure*}

\noindent\textbf{Domain Specialization of Images in Experts:} Figure \ref{fig:routing} visualizes the activation states of MoE experts during inference. We sample 200 data points across different modalities and datasets, identifying the top-1 expert with the most activations in each MoE layer. The visualization shows domain specialization for different input image modalities: Expert 1 and Expert 2 for CT, Expert 2 and Expert 3 for MRI, Expert 4 for Pathology, and Expert 1 for X-Ray. This specialization, due to our router and meta experts, enhances MoE performance by encouraging each expert to focus on specific modalities and collaborate with the meta expert for global information. However, visualizing expert activations for four modalities handled by traditional routers in each MoE layer reveals fused patterns, resulting in weaker interpretability and performance.

\noindent\textbf{Cost Analysis:} Table \ref{tab:cost_analysis} illustrates the cost efficiency of our models compared to LLaVA-Med. Particularly, Med-LoRAMoE models show significant reductions in training GPU memory usage and inference time. For example, Med-LoRAMoE (StableLM) requires only 8GB of GPU memory and 3 seconds for inference, demonstrating high efficiency for deployment, and making our approach more appealing to practical resource-constrained clinical settings.

\begin{table}[ht!]
\centering
\scriptsize
\setlength{\tabcolsep}{1pt} 
\renewcommand{\arraystretch}{1.2} 
\resizebox{0.5\textwidth}{!}{ 
\begin{tabular}{l|c|c|c|c}
\hline
\textbf{Model} & \textbf{Model Size} & \textbf{Training GPU} & \textbf{Inference Time} & \textbf{Load Memory} \\
\hline
LLaVA-Med & 7B & >24GB & 5s & 20GB \\
Med-MoE (Phi2) & 3.6B & 23GB & 3s & 13.4GB \\
Med-MoE (StableLM) & 2.0B & 12.5GB & 3s & 10.5GB \\
Med-LoRAMoE (Phi2) & 3.6B & 13.5GB & 3s & 13.7GB \\
Med-LoRAMoE (StableLM) & 2.0B & 8GB & 3s & 10.8GB \\
\hline
\end{tabular}
}
\caption{Cost efficiency comparison of different models.}
\label{tab:cost_analysis}
\end{table}

\begin{table}[ht!]
\centering
\scriptsize
\setlength{\tabcolsep}{10pt} 
\renewcommand{\arraystretch}{1.2} 
\resizebox{0.5\textwidth}{!}{ 
\begin{tabular}{l|c|c}
\hline
\textbf{Top-k (Experts=4)} & \textbf{1 Expert} & \textbf{2 Experts(Ours)} \\
\hline
{VQA-RAD (Open)} & 46.7 & 47.2 \textbf{(+0.5)} \\
{VQA-RAD (Closed)} & 83.4 & 83.8 \textbf{(+0.4)} \\
{SLAKE (Open)} & 82.1 & 82.3 \textbf{(+0.2)} \\
{SLAKE (Closed)} & 83.2 & 84.9 \textbf{(+1.7)} \\
{PathVQA (Open)} & 33.9 & 34.1 \textbf{(+0.2)} \\
{PathVQA (Closed)} & 90.9 & 91.8 \textbf{(+0.9)} \\
{Time} & 7h & 8h \\
\hline
\end{tabular}
}
\caption{Performance with varying activated experts}
\label{table:experts}
\end{table}

\noindent\textbf{Effect of the Number of Activated Experts: } The router's top-$k$ selection is typically 1 or 2 in many existing MoE works, as more would negate sparsity benefits and increase memory overhead. We evaluate performance with different numbers of activated experts, as shown in Table \ref{table:experts}. Ultimately, we chose 4 experts with top-2 activations for balanced performance and overhead.

\begin{table}[ht!]
\centering
\setlength{\tabcolsep}{7pt} 
\renewcommand{\arraystretch}{1.2} 
\resizebox{0.5\textwidth}{!}{ 
\begin{tabular}{l|c|c|c}
\hline
\textbf{Experts (Top-k=2)} & \textbf{2 Experts} & \textbf{4 Experts(Ours)} & \textbf{6 Experts} \\
\hline
{VQA-RAD (Open)} & 47.03 \textbf{(-0.8)} & 47.83 & 48.03 \textbf{(+0.2)} \\
{VQA-RAD (Closed)} & 77.54 \textbf{(-1.5)} & 79.04 & 78.84 \textbf{(-0.2)} \\
{SLAKE (Open)} & 81.52 \textbf{(-0.6)} & 82.12 & 82.42 \textbf{(+0.3)} \\
{SLAKE (Closed)} & 81.45 \textbf{(-1.0)} & 82.45 & 82.65 \textbf{(+0.2)} \\
{PathVQA (Open)} & 32.78 \textbf{(-0.5)} & 33.28 & 33.58 \textbf{(+0.3)} \\
{PathVQA (Closed)} & 90.60 \textbf{(-0.2)} & 90.80 & 91.10 \textbf{(+0.3)} \\
{Time} & 6h & 8h & 11h \\
\hline
\end{tabular}
}
\caption{Performance with varied expert number}
\label{table:numberofexperts}
\end{table}

\noindent\textbf{Effect of the Number of Experts: } In MoE applications, choosing the right number of experts is crucial for balancing performance and computational efficiency. We evaluate configurations with 2, 4, and 6 experts, each with the top-2 activations. Results in Table \ref{table:numberofexperts} show that using 4 experts achieves the best balance between performance and cost efficiency. While increasing to 6 experts offers slight performance gains, it significantly increases computational time and memory usage. On the other hand, reducing the number to 2 experts leads to decreased performance across all tasks. 

\section{Related Work}
\noindent\textbf{Medical MLLMs}
Advancements in Medical MLLMs, such as Med-Flamingo~\citep{moor2023med}, Med-PaLM M~\citep{singhal2023towards}, and LLaVA-Med~\citep{li2024llava}, have significantly impacted medical diagnostics and patient care, building on general AI models like ChatGPT ~\citep{chatgpt} and GPT-4~\citep{gpt4}. These models enhance few-shot learning, medical question answering, and conversational AI, demonstrating the potential of specialized MLLMs in healthcare. Biomedical chatbots like ChatDoctor ~\citep{yunxiang2023chatdoctor} and Visual Med-Alpaca highlight the benefits of domain-specific fine-tuning. However, their application in resource-constrained hospital settings remains underexplored, emphasizing the need for cost-efficient MLLMs in clinical contexts.

\noindent\textbf{MoE in MLLMs}
MoE in MLLMs addresses task conflicts in multi-task learning and offers a cost-efficient scaling method. The first approach~\citep{xu2024meteora,chen2024llava,gou2023mixture} uses Top-1 activation to assign different tasks to different experts, avoiding performance degradation from task data conflicts but overlooking modal biases within the same task type. The second approach~\citep{lin2024moe,li2024uni,lee2024moai,liu2024mixture,dai2024deepseekmoe} replaces FFN layers in LLMs with MoE structures using multiple expert activations, achieving improvements with minimal additional parameters. However, visualizing the expert activations in routers shows these methods often fail to specialize experts effectively, limiting interpretability and performance with diverse data modalities~\citep{fan2024towards}. A specialized MoE architecture tailored to the medical domain is needed to leverage modality-specific information and improve performance with a smaller LLM backbone.

\section{Conclusion}
We have introduced Med-MoE, a lightweight framework for multimodal medical tasks, addressing both discriminative and generative needs. Optimized for resource-constrained environments, Med-MoE involves aligning medical images with language model tokens, task-specific instruction tuning, and domain-specific expert fine-tuning. Our approach reduces activated parameters while maintaining or surpassing state-of-the-art performance. Our experiments on VQA-RAD, SLAKE, and Path-VQA validate Med-MoE's effectiveness and efficiency. This model offers a practical solution for deploying advanced medical AI in diverse and resource-limited clinical settings.

\section{Discussion and Limitations}

Our work primarily sought to develop a smaller, more cost-efficient Multimodal Large Language Model (MLLM) for the medical field, diverging from the current trend focused on creating larger and more robust models. We posit that in practical applications, especially in resource-constrained environments like mobile devices, smaller models could be more advantageous. This approach not only addresses the practical limitations of deploying large-scale models in routine clinical settings but also explores the feasibility of using leaner models without compromising on performance, fostering broader accessibility and application.

However, our approach faces several limitations. First, there is a notable scarcity of training data in the medical domain, largely due to the sensitivity and privacy concerns associated with medical data. Generating synthetic data through methods like those used for GPT-4V can be problematic in this context, and many datasets require labor-intensive manual annotations by medical professionals. This is both costly and limits the scalability of data generation efforts. As illustrated in Supplementary Figure \ref{fig:sup3}, our model occasionally fails, particularly with more complex open-ended questions that demand precise medical knowledge.

Furthermore, the inherent requirement for medical applications to provide trustworthy explanations and confidence scores poses another challenge. Ensuring that the model outputs are not only accurate but also accompanied by reliable justifications is crucial, especially in a field where decisions have significant health implications. This necessity heightens the importance of building a trustworthy MLLM that can articulate its reasoning processes clearly and provide confidence levels, thereby enhancing the reliability and safety of AI applications in healthcare.

\section*{Acknowledgments}
\bibliography{custom}

\appendix

\section{Appendix}
\label{sec:appendix}
\subsection*{Calculation Formulas for Open Setting Metrics}
\label{sec:gongshi}
\subsubsection*{1. Recall}
Recall is calculated using the following formula:
\begin{align}
\text{Recall} = \frac{TP}{TP + FN}
\end{align}
where \(TP\) (true positives) is the number of words in both the candidate and the reference, and \(FN\) (false negatives) is the number of words in the reference but not in the candidate.

\subsubsection*{2. Exact Match Score}
Exact Match Score is calculated using the following formula:
\begin{align}
\text{EMS} = \frac{\text{Number of matching words}}{\text{Total number of candidate words}}
\end{align}
This formula calculates the ratio of the number of matching words in the candidate and the reference to the total number of words in the candidate.

\subsubsection*{3. BLEU Score}
The BLEU Score is calculated using the following steps:
\begin{itemize}
    \item Calculate the modified precision \(p_n\) for each n-gram up to \(n\):
    \begin{equation}
    \resizebox{\linewidth}{!}{$
    p_n = \frac{\sum_{C \in \text{Candidates}} \sum_{ng \in C} \min(\text{Count}(ng), \text{Count}_{\text{max}}(ng))}{\sum_{C \in \text{Candidates}} \sum_{ng \in C} \text{Count}(ng)}
    $}
    \end{equation}
    where \(ng\) is the n-gram, \(\text{Count}(ng)\) is its count in the candidate, and \(\text{Count}_{\text{max}}(ng)\) is its maximum count in the reference.
    
    \item Calculate the brevity penalty (BP):
    \begin{equation}
    \text{BP} = 
    \begin{cases} 
    1 & \text{if } c > r \\
    e^{(1 - \frac{r}{c})} & \text{if } c \leq r
    \end{cases}
    \end{equation}
    where \(c\) is the length of the candidate sentence and \(r\) is the length of the reference sentence.
    
    \item Combine the modified precision and brevity penalty to compute the BLEU score:
    \begin{equation}
    \text{BLEU} = \text{BP} \cdot \exp\left( \sum_{i=1}^{n} w_i \log p_i \right)
    \end{equation}
    where \(w_i\) are the weights assigned to each n-gram precision.
\end{itemize}
\subsection{Dataset information}
VQA-RAD~\citep{lau2018dataset} contains 3,515 QA pairs and 315 radiology images, with questions covering 11 categories and a mix of closed-ended and open-ended types. SLAKE~\citep{liu2021slake} comprises 642 radiology images and over 7,000 QA pairs, including segmentation masks and object detection bounding boxes. PathVQA~\citep{he2020pathvqa} includes 4,998 pathology images with 32,799 QA pairs, focusing on aspects like location, shape, color, and appearance, categorized into open-ended and closed-ended types. The PneumoniaMNIST dataset focuses on pediatric chest radiographs for binary classification of pneumonia versus normal, using 4,708 training and 624 test images. OrganCMNIST ~\citep{yang2023medmnist} classifies 11 human body organs with 12,975 training and 8,216 testing images. To ensure a fair comparison with LLaVA-Med, we did not use the additional image classification training datasets when evaluating VQA. For evaluation, we use test sets from these widely recognized medical VQA datasets and additionally assess classification performance. 

\begin{table*}[ht!]
\centering
\begin{tabular}{l|c|c|c}
\toprule
\textbf{Config} & \textbf{Stage I} & \textbf{Stage II} & \textbf{Stage III} \\
\midrule
Deepspeed & \multicolumn{3}{c}{Zero2, Zero2, Zero2 offload} \\
Image encoder & \multicolumn{3}{c}{CLIP-Large} \\
Feature select layer & \multicolumn{3}{c}{-2} \\
Image projector & \multicolumn{3}{c}{2 Linear layers with GeLU} \\
Epoch (same as LLaVA-Med) & 1 & 3 & 9 \\
Learning rate & 1e-3 & 2e-5 & 2e-5 \\
Learning rate schedule & \multicolumn{3}{c}{Cosine} \\
Weight decay & \multicolumn{3}{c}{0.0} \\
Text max length & \multicolumn{3}{c}{2048} \\
Batch size per GPU & \multicolumn{3}{c}{2} \\
GPU & \multicolumn{3}{c}{8 × 3090-24G} \\
Precision & \multicolumn{3}{c}{Bf16} \\
\bottomrule
\end{tabular}
\caption{Our experimental hyperparameters}
\label{tab:setting}
\end{table*}

\begin{figure}[!ht]
    \centering
    \begin{adjustbox}{width=\linewidth, keepaspectratio}
        \includegraphics{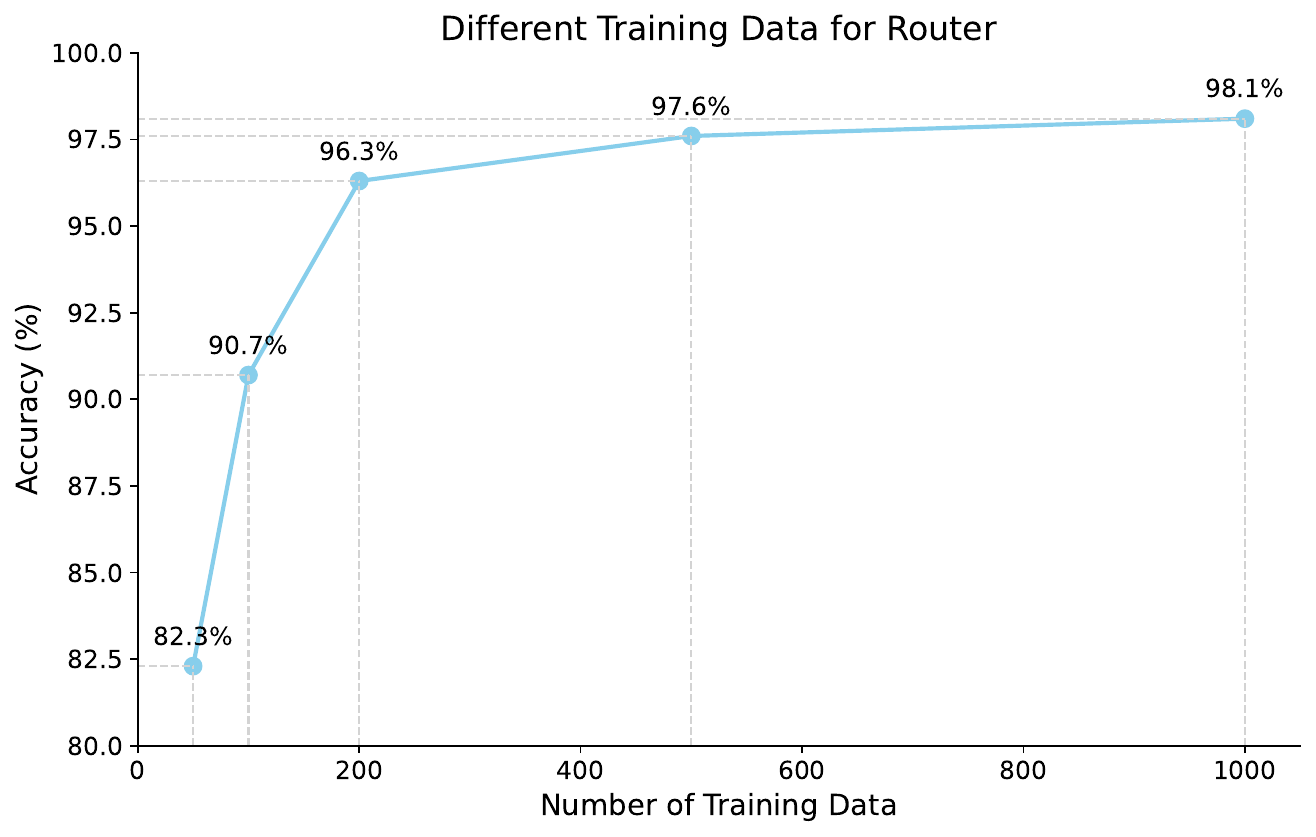}
    \end{adjustbox}
    \caption{Performance of router predictions with different domain-labeled training data}
    \label{fig:data_router}
\end{figure}

\begin{table*}[!ht]
\centering
\small
\setlength{\tabcolsep}{15pt} 
\renewcommand{\arraystretch}{1.5} 
\resizebox{\textwidth}{!}{%
\begin{tabularx}{\textwidth}{|c|X|c|}
\hline
\textbf{Stage} & \textbf{Data Source} & \textbf{Sample Size} \\
\hline
Stage 1 & llava\_med\_alignment\_500k.json & 500K \\
\hline
Stage 2 & instruct\_60k\_inline\_mention & 60K \\
\hline
\multirow{2}{*}{Stage 3} & \textbf{VQA:} RAD-VQA, SLAKE, Path-VQA: 27K & \multirow{2}{*}{44K} \\
\cline{2-2}
& \textbf{Classification:} PneumoniaMNIST, OrganCMNIST: 17K & \\
\hline
\end{tabularx}
}
\caption{Summary of Data Utilized Across Training Stages}
\label{tab:data_summary}
\end{table*}

\begin{table*}[!ht]
\centering
\resizebox{\textwidth}{!}{
\begin{tabular}{l|c|c|c|c|c|c|c|c|c|c|c}
\hline
\textbf{Name} & \textbf{Experts} & \textbf{Activated Experts} & \textbf{MoE Layers} & \textbf{Embedding} & \textbf{Width} & \textbf{Layers} & \textbf{FFN} & \textbf{FFN Factor} & \textbf{Heads} & \textbf{Activated Param} & \textbf{Total Param} \\
\hline
StableLM-1.6B & - & - & - & 100352 & 2560 & 32 & 10240 & 2 & 32 & 1.6B & 1.6B \\
\rowcolor[HTML]{CBC5D3} Med-MoE (StableLM-4x1.6B) & 4 & 2 & 16 & 100352 & 2560 & 32 & 10240 & 2 & 32 & 2.0B & 2.9B \\
\hline
Phi2-2.7B & - & - & - & 51200 & 2560 & 32 & 10240 & 2 & 32 & 2.7B & 2.7B \\
\rowcolor[HTML]{CBC5D3} Med-MoE (Phi2-4x2.7B) & 4 & 2 & 16 & 51200 & 2560 & 32 & 10240 & 2 & 32 & 3.6B & 5.3B \\
\hline
\end{tabular}
}
\caption{Comparison of different models in terms of various parameters.}
\label{tab:model_arichi}
\end{table*}

\begin{figure*}[!ht]
    \centering
    \includegraphics[width=1\linewidth]{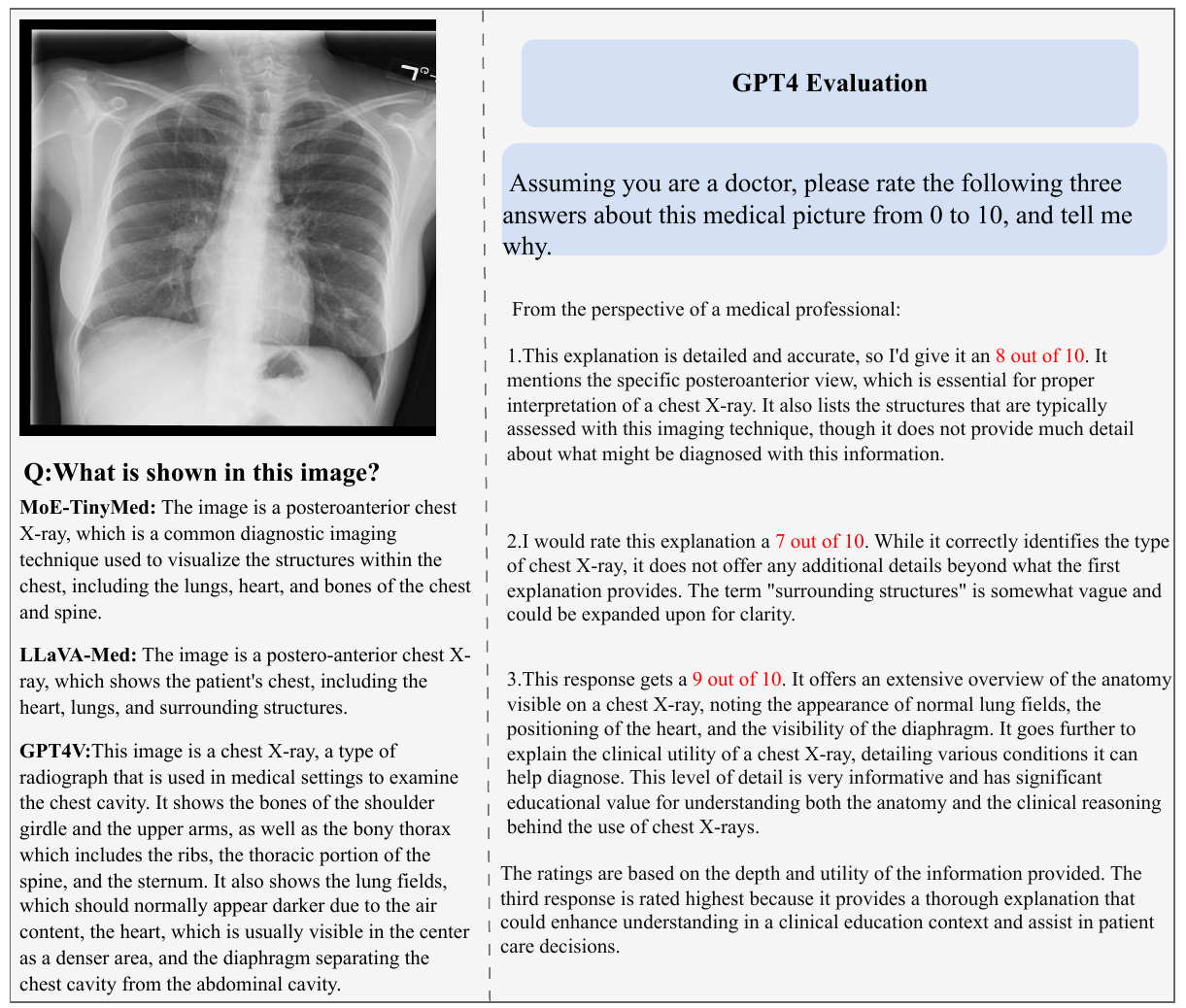}
    \caption{An example showcasing our method's ability to answer medical imaging questions with performance nearing or even surpassing that of LLaVA-Med under GPT-4V evaluation.}
    \label{fig:sup1}
\end{figure*}

\begin{figure*}[!ht]
    \centering
    \includegraphics[width=1\linewidth]{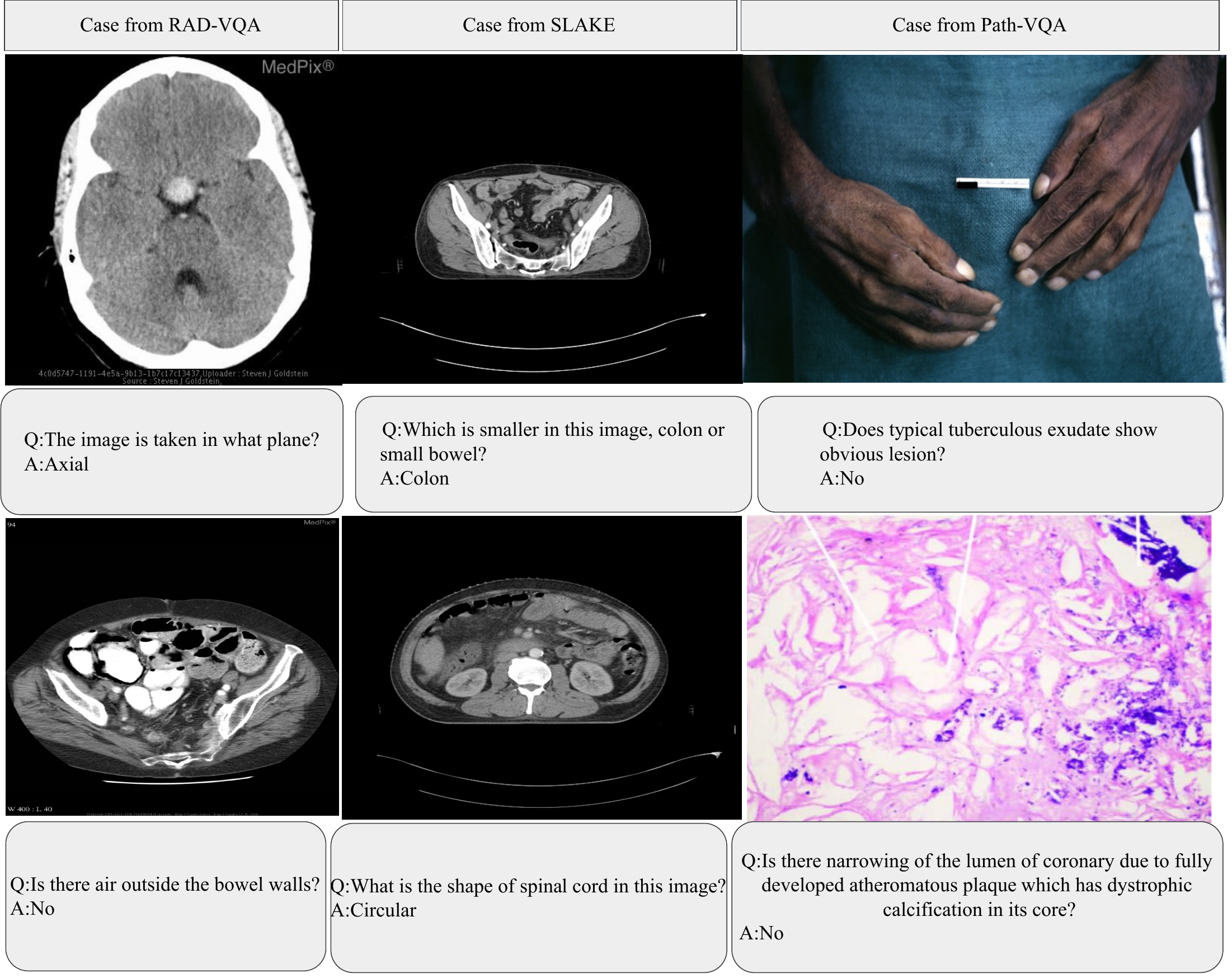}
    \caption{More Medical VQA cases from VQA-RAD, SLAKE, and Path-VQA. our Med-MoE generates expected responses for medical image queries.}
    \label{fig:sup2}
\end{figure*}

\begin{figure*}[!ht]
    \centering
    \includegraphics[width=1\linewidth]{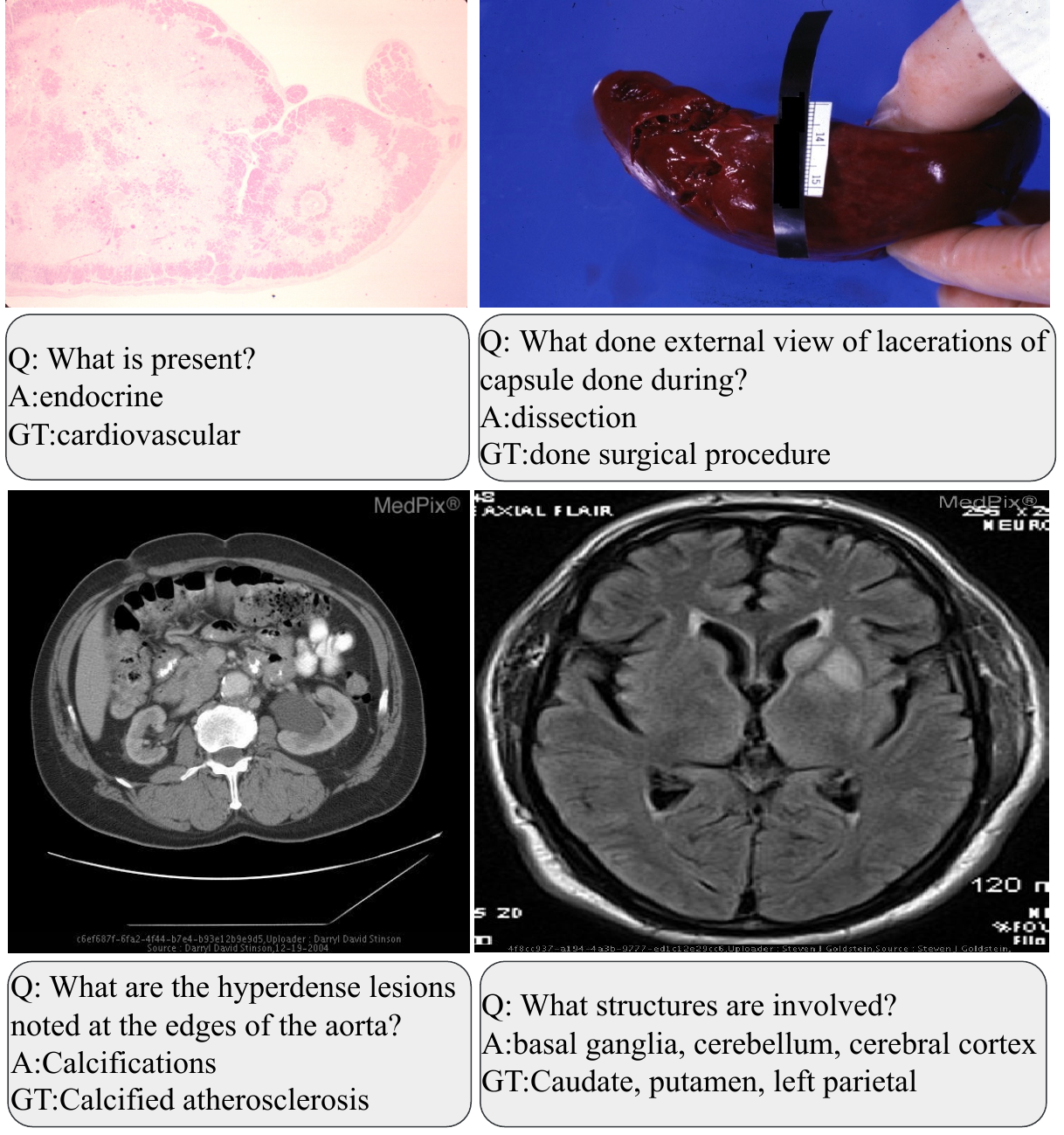}
    \caption{Incorrect cases: Med-VQA examples in OPEN setting requiring precise and specialized medical knowledge.}
    \label{fig:sup3}
\end{figure*}

\begin{figure*}[!ht]
    \centering
    \includegraphics[width=1\linewidth]{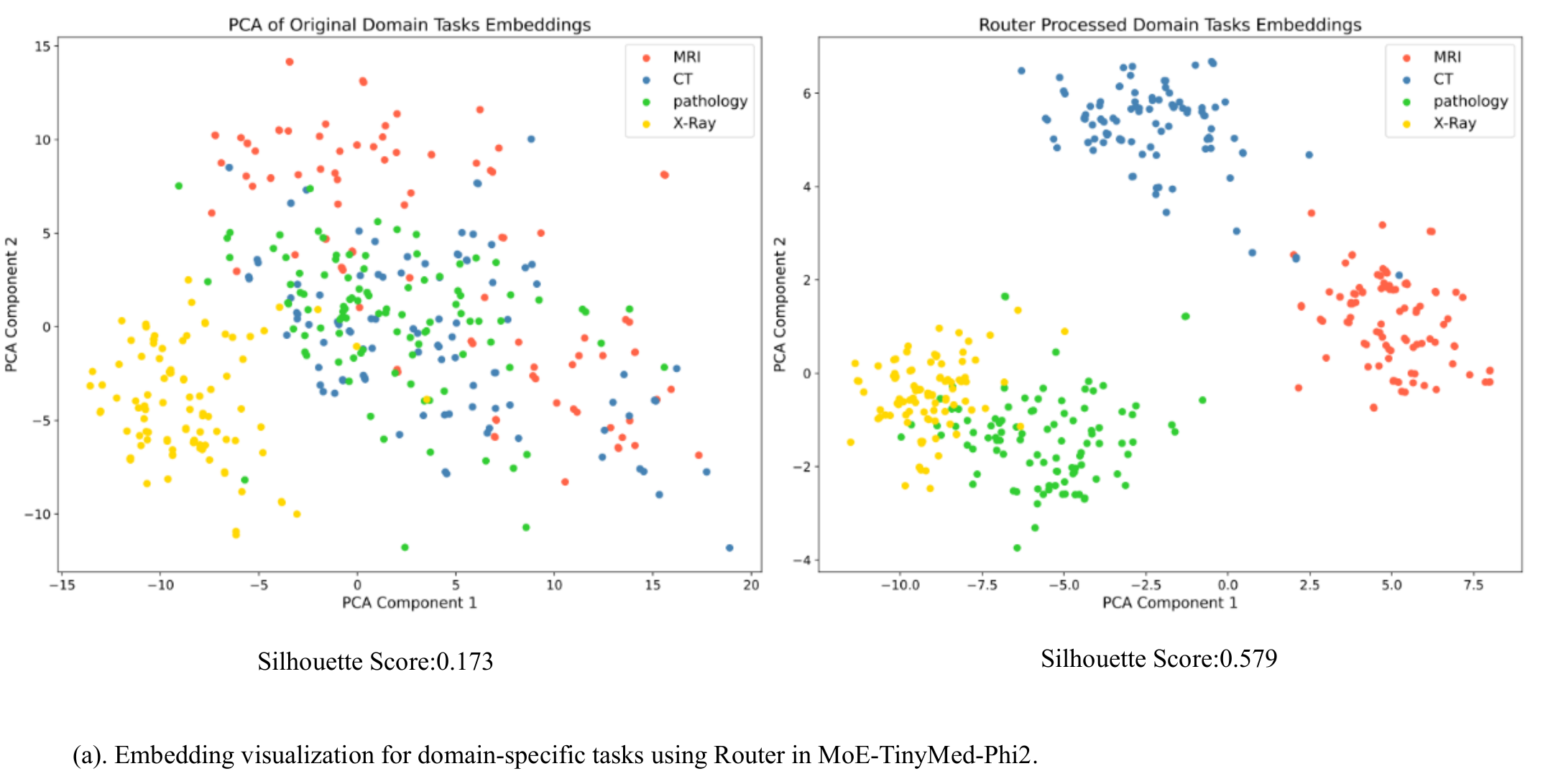}
\caption{Experts Routing Visualization On Phi2}
\label{fig:router_sup}
\end{figure*}

\begin{figure*}[!t]
    \centering
    \includegraphics[width=1\linewidth]{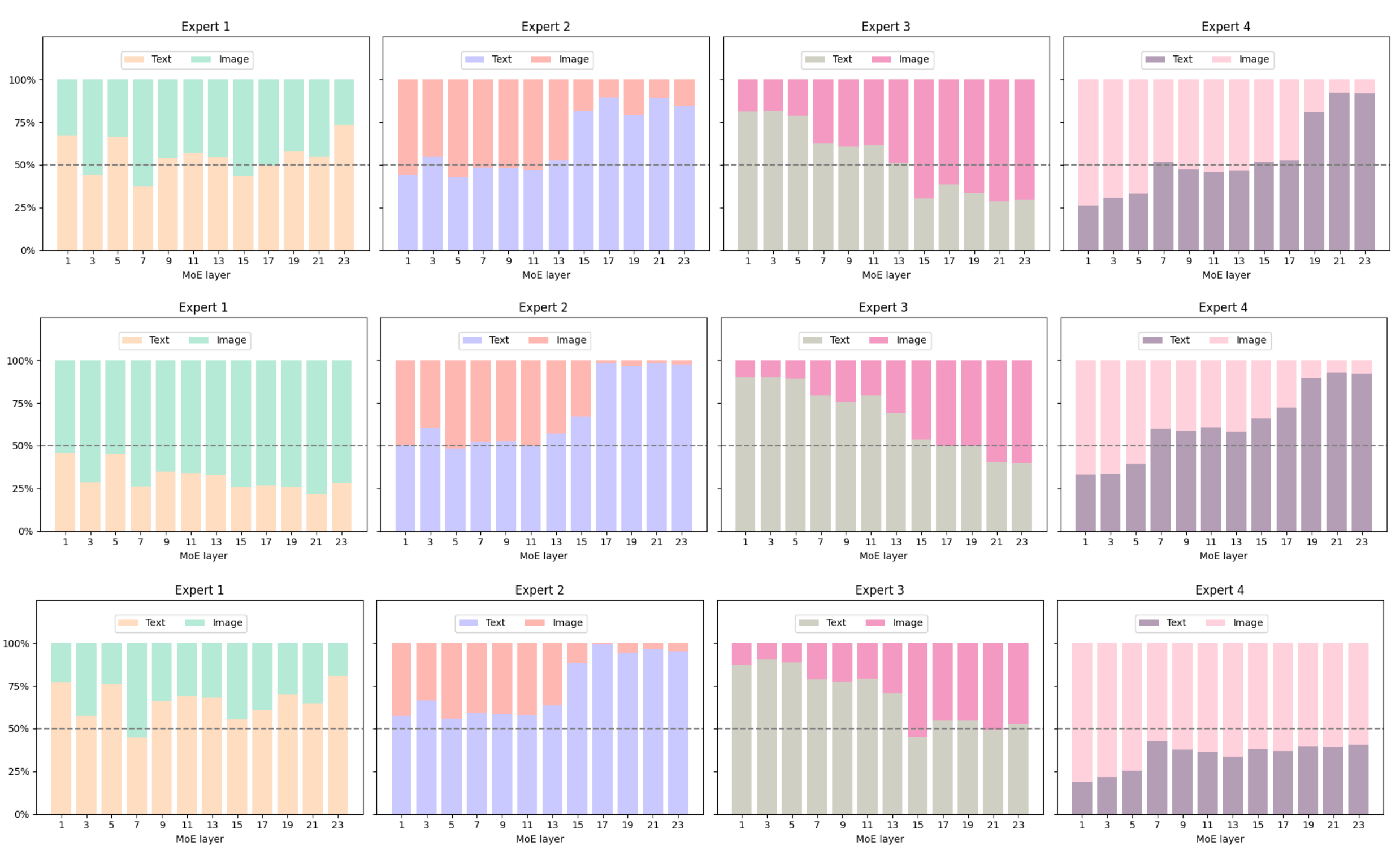}
\caption{The activation proportions for text and image processing in other modalities}
\label{fig:other_image_token}
\end{figure*}

\end{document}